\documentclass[sigconf]{acmart}
\usepackage{balance}
\AtBeginDocument{%
  \providecommand\BibTeX{{%
    \normalfont B\kern-0.5em{\scshape i\kern-0.25em b}\kern-0.8em\TeX}}}

\copyrightyear{2022}
\acmYear{2022}
\setcopyright{rightsretained}

\acmConference[SIGIR '22]{Proceedings of the 45th International ACM SIGIR Conference on Research and Development in Information Retrieval}{July 11--15, 2022}{Madrid, Spain.}
\acmBooktitle{Proceedings of the 45th Int'l ACM SIGIR Conference on Research and Development in Information Retrieval (SIGIR '22), July 11--15, 2022, Madrid, Spain}
\acmDOI{10.1145/3477495.3531875}
\acmISBN{978-1-4503-8732-3/22/07}

\usepackage{etoolbox}
\makeatletter
\patchcmd{\maketitle}{\@copyrightpermission}{
   \begin{minipage}{0.3\columnwidth}
     \href{http://creativecommons.org/licenses/by/4.0/}{\includegraphics[width=0.90\textwidth]{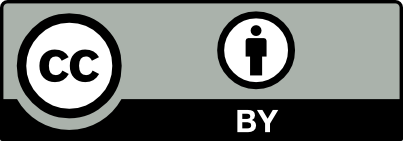}}
   \end{minipage}\hfill
   \begin{minipage}{0.7\columnwidth}
     \href{http://creativecommons.org/licenses/by/4.0/}{This work is licensed under a Creative Commons Attribution International 4.0 License.}
   \end{minipage}

   \vspace{5pt}
}{}{}

\makeatother
 
\usepackage{graphicx}
\usepackage{caption}
\usepackage{cprotect}
\usepackage{subcaption}
\usepackage{booktabs}
\usepackage{multirow}
\usepackage{stfloats}
\begin{document}
\fancyhead{}
%%
%% The "title" command has an optional parameter,
%% allowing the author to define a "short title" to be used in page headers.
% \title{Spam Detection at Scale with Relational Graph Neural Model}
\title{Modeling User Behavior With Interaction Networks for Spam Detection}

%%
%% The "author" command and its associated commands are used to define
%% the authors and their affiliations.
%% Of note is the shared affiliation of the first two authors, and the
%% "authornote" and "authornotemark" commands
%% used to denote shared contribution to the research.

% \author{Prabhat Agarwal, Manisha Srivastava, Vishwakarma Singh, Chuck Rosenberg}
% \email{{pagarwal,manishasrivastava,vishwakarmasingh,chuck}@pinterest.com}
% \affiliation{%
%   \institution{Pinterest}
%   \country{USA}
% }
\author{Prabhat Agarwal}
\email{pagarwal@pinterest.com}
\orcid{0000-0002-3826-0858}
\affiliation{%
  \institution{Pinterest}
  \country{USA}
}

\author{Manisha Srivastava}
\email{manishasrivastava@pinterest.com}
\affiliation{%
  \institution{Pinterest}
  \country{USA}
}

\author{Vishwakarma Singh}
\email{vishwakarmasingh@pinterest.com}
\orcid{0000-0002-2991-2236}
\affiliation{%
  \institution{Pinterest}
  \country{USA}
}

\author{Charles Rosenberg}
\email{crosenberg@pinterest.com}
\affiliation{%
  \institution{Pinterest}
  \country{USA}
}

%%
%% By default, the full list of authors will be used in the page
%% headers. Often, this list is too long, and will overlap
%% other information printed in the page headers. This command allows
%% the author to define a more concise list
%% of authors' names for this purpose.
\renewcommand{\shortauthors}{Agarwal, et al.}

\newcommand{\vishwakarma}[1]{{{\textcolor{red}{[Vishwakarma: #1]}}}}
\newcommand{\prabhat}[1]{{{\textcolor{blue}{[Prabhat: #1]}}}}
\newcommand{\manisha}[1]{{{\textcolor{green}{[Manisha: #1]}}}}

\newcommand{\modelname}{{\textit{SEINE }}}
\newcommand{\modelnamenospace}{{\textit{SEINE}}}

\setlength\intextsep{2pt}
%%
%% The abstract is a short summary of the work to be presented in the
%% article.
\begin{abstract}
Spam is a serious problem plaguing web-scale digital platforms which facilitate user content creation and distribution. It compromises platform's integrity, performance of services like recommendation and search, and overall business. Spammers engage in a variety of abusive and evasive behavior which are distinct from non-spammers. Users' complex behavior can be well represented by a heterogeneous graph rich with node and edge attributes. Learning to identify spammers in such a graph for a web-scale platform is challenging because of its structural complexity and size. In this paper, we propose \modelname (\textbf{S}pam D\textbf{E}tection using \textbf{I}nteraction \textbf{NE}tworks), a spam detection model over a novel graph framework. Our graph simultaneously captures rich users' details and behavior and enables learning on a billion-scale graph. Our model considers neighborhood along with edge types and attributes, allowing it to capture a wide range of spammers. \modelnamenospace, trained on a real dataset of tens of millions of nodes and billions of edges, achieves a high performance of $\mathbf{80\%}$ recall with $\mathbf{1\%}$ false positive rate. \modelname achieves comparable performance to the state-of-the-art techniques on a public dataset while being pragmatic to be used in a large-scale production system.

%Most of the prior work has either addressed a specific kind of abuse or captured only partial users data resulting into low coverage of spammers, or do not scale to large graphs. 

%which in turn harms users' trust 
%\textcolor{red}{We learn the model on a real-data graph of millions of nodes and hundreds of millions of edges and infer hundreds of millions of users. Our technique yields production level performance with \{recall percentage\} recall with a false positive rate of \{FPR\%\}. It outperforms strong baselines and prior techniques by \{performance improvement \%\}}.
\end{abstract}

%%
%% The code below is generated by the tool at http://dl.acm.org/ccs.cfm.
%% Please copy and paste the code instead of the example below.
%%


%%
%% Keywords. The author(s) should pick words that accurately describe
%% the work being presented. Separate the keywords with commas.

%% A "teaser" image appears between the author and affiliation
%% information and the body of the document, and typically spans the
%% page.
% \begin{teaserfigure}
%   \includegraphics[width=\textwidth]{figures/master-fig.png}
%   \caption{Wide-range of spam behaviors and their translation to heterogeneous graph.}
%   \Description{Wide-range of spam behaviors and their translation to heterogeneous graph}
%   \label{fig:master-fig}
% \end{teaserfigure}

%%
%% This command processes the author and affiliation and title
%% information and builds the first part of the formatted document.
\maketitle

\section{Introduction}
Online platforms (Pinterest, Facebook, Snap, Youtube, etc.) are the go-to place for people to share content and information, interact, and drive influence. These platforms also give earning opportunities to users based on the engagement generated by their content. Some bad actors with malicious intent abuse these platforms, either individually or in a group, for unfair gains. Spam \cite{webspamreview:2012, spambehavanalysis:2012} is a common form of abuse that compromises platform's integrity, performance of services like recommendation and search, and users' trust. It is an umbrella term for a wide-class of abuses: posting spam content ~\cite{spam-types-microblogs:2013}; artificial engagement boosting \cite{fake-engagement:2016}; evading the system; diverting traffic through links; inorganic follow \cite{followspam:2010, sn-maliciousgroups:2009}; and collusion \cite{sliceNDice}. Addressing spam by abuse-specific solutions is neither practical nor cost-effective in the industry because of their wide range and emergence of new spam vectors with product evolution. To make things worse, abusers also display adversarial nature by resorting to innovative evasive techniques against the deployed solutions. Designing a holistic solution to identify spam users (spammers), the root of the problem, yields the best strategy to effectively fight spam. Hence, we address the problem of identifying spammers on a web-scale platform in this paper.

\begin{figure*}[tb]
     \centering
     \begin{subfigure}[b]{0.49\textwidth}
         \centering
         \includegraphics[width=\textwidth]{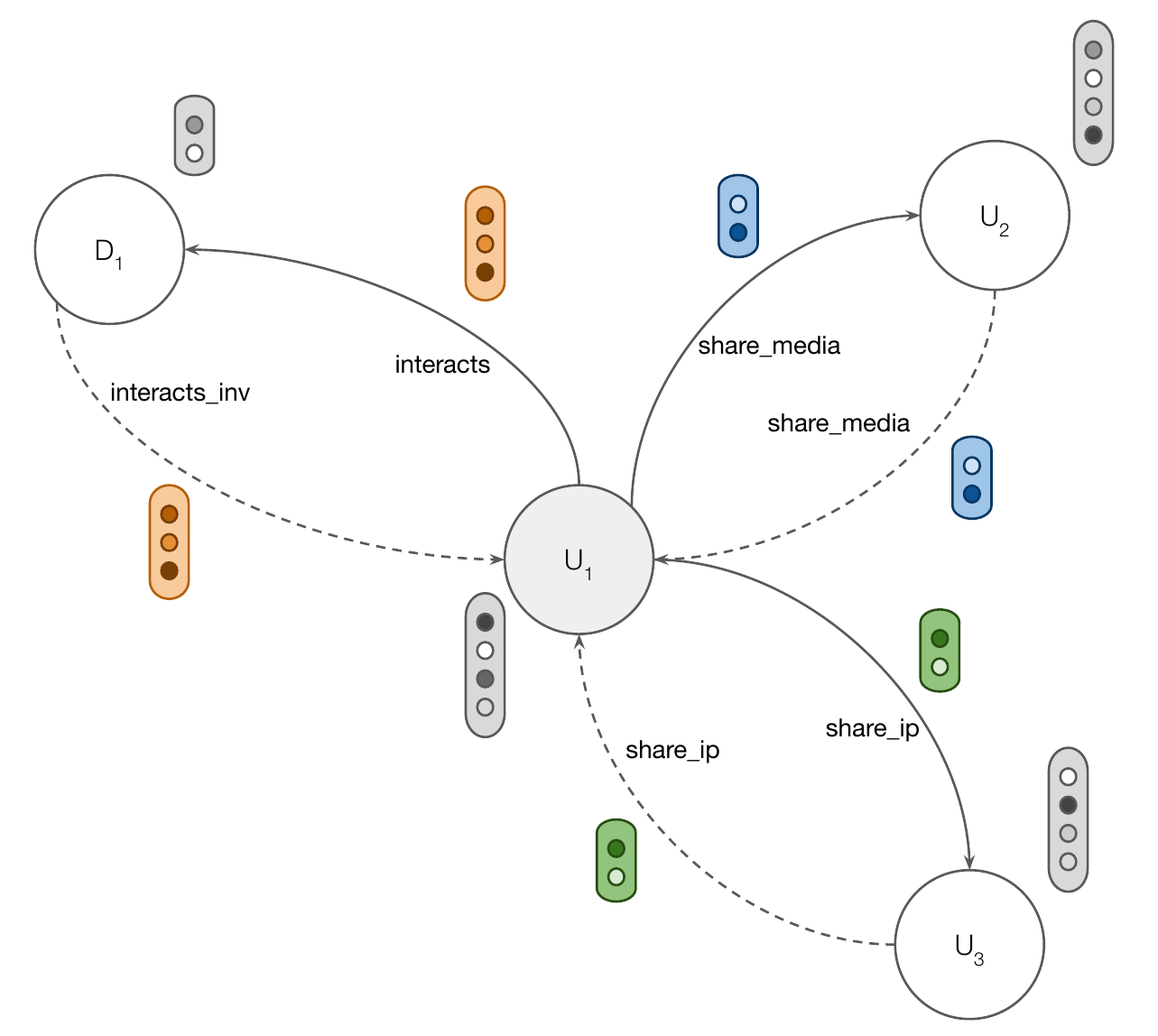}
         \caption{User Interaction Graph}
     \end{subfigure}
     \hfill
     \begin{subfigure}[b]{0.49\textwidth}
         \centering
         \includegraphics[width=\textwidth]{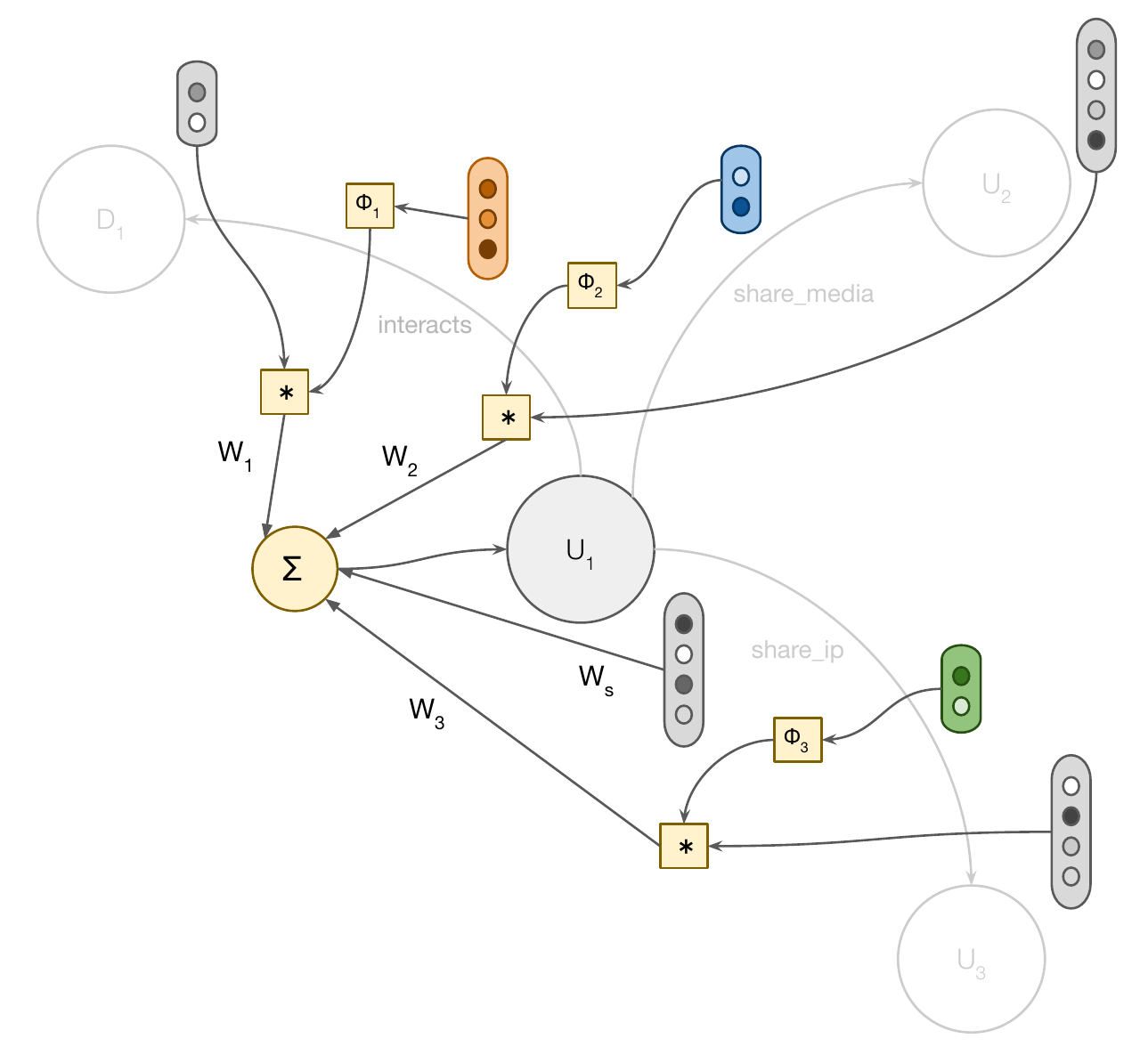}
         \caption{\modelname update}
     \end{subfigure}
        \caption{A schematic overview of \modelnamenospace. The left figure (a) is a subgraph of the user interaction graph showing the neighbors (users and domains) of a user $U_1$ with the node and edge features.  The right figure (b) describes the convolution for user $U_1$ at a given layer in the  user-interaction graph. \modelname first calculates edge weights $w_e$ using a relation specific function $\phi_r$ and the edge features. Then it aggregates node embeddings of the neighbors using relation specific filters $W_r$ and edge weights $w_e$. Finally, a sum of the neighbors aggregate and $U_1$'s embedding is transformed to produce the updated embedding for $U_1$. Please refer to Section~\ref{model} for details.}
         \label{fig:model_overview}
    \vspace{-0.9em}
\end{figure*}

Spammers exhibit different patterns from genuine users (non-spammers) in their profile and behavior which is evident from our analysis and other prior works \cite{Oddball:2010, Liu2018HeterogeneousGN, spambehavanalysis:2012}. We broadly define a user’s behavior by its types and rate of activities and interaction with other users and entities (e.g., media, IP, weblinks) on the platform. This behavior can only be truly captured by a heterogeneous graph where nodes represent users and entities, edges represent the interactions of users with other users or entities, node attributes represent details of a user or an entity, and edge attributes represent interaction type and other information. Detecting spammers on web-scale platforms using such a graph is challenging because of its large size and structural complexity. Prior works \cite{graphmethodsurvey:2015, graphanomaly1:2003, graphanomaly2:2005, sliceNDice} for abuse detection using graphs either address a specific kind of abuse or use a smaller setting that does not scale to a large graph, or does not capture the spectrum of behavior that enables a solution with a wide abuse coverage. Recently graph neural networks (GNNs) \cite{zeng2021rlc, peng2021reinforced, Liu2018HeterogeneousGN} have been explored to detect abuse on a platform but have similar shortcomings as others or lack pragmatism.
In this paper, we provide analysis on a large real dataset from Pinterest to get insights into differentiating patterns of spammers from non-spammers. We use these insights to design a new kind of heterogeneous graph and learn a model on such a graph to detect spammers on a web-scale platform. Our major contributions are:
\vspace{-\topsep}
\begin{itemize}
 \item A user-entity graph that uses edge types and attributes along with node attributes to capture more behavior than considered in prior works. It also selectively transforms some of the user-entity interactions into user-user edges to reduce the size of the graph and enables learning a model on a large-scale graph.
 \item A graph neural network model \modelname (\textbf{S}pam D\textbf{E}tection using \textbf{I}nteraction \textbf{NE}tworks) that considers neighborhood along with edge types and attributes for differentiating spammers from non-spammers.
 \item An extensive empirical evaluation on a large graph from Pinterest which establishes that \modelname has $\mathbf{40\%}$ better performance over strong ablation baselines. \modelname achieves comparable performance to the state-of-the-art methods on a public dataset while being simple and pragmatic to be used in a large-scale production system.
\end{itemize}
\vspace{-\topsep}
We discuss related work in Section \ref{related work},  user interaction graph in Section \ref{sec:user_interaction_graph}, problem definition in Section \ref{problem_defintion}, model  in Section \ref{model}, empirical evaluations in Section \ref{experiments}, and conclusions in Section \ref{conclusion}.

\section{Related Work}
\label{related work}
Researchers have proposed a variety of spam detection techniques using both content and behavior features for various domains: web \cite{webspamreview:2012, trustrank:2004}, email \cite{BotGraph:2009, emailspamreview1:2007, emailspamreview2:2008}, microblogging \cite{followspam:2010, microblogspam:2013}, social networks \cite{socialnetworkspam:2012, sn-maliciousgroups:2009}, and reviews \cite{reviewspam:2007, reviewspamsurvey:2015}. Most of these techniques apply classical machine learning, use human-created features represented in tabular data format on a small-scale dataset, and only partially capture the complex behavior of spammers. Graph representation \cite{graphmethodsurvey:2015} has been explored for anomaly detection and then applied for spam. \cite {graphanomaly1:2003} proposes an unsupervised technique to find anomalous subgraphs, \cite{graphanomaly2:2005} uses density and grid-based clustering to find anomalous clusters, and \cite{sn-maliciousgroups:2009} clusters users based on the similarity of their activities over a sustained time. Prior works have also explored mining techniques \cite{Eigenspokes:2010, Fraudar:2016, crossasscoaition:2004, inf-the-min-len:2003, sliceNDice} for detecting anomalous groups.  Most of these graph techniques either use a homogeneous graph, or relatively small graphs and can not be scaled to large graphs, or detect only clusters of colluding spammers and do not address individual spammers. Recent advances in Graph Neural Networks \cite{kipf2016semi, hamilton2017inductive, peng2021reinforced, rGCN, vashishth2019composition} and transformer architectures \cite{vaswani2017attention} have received some attention for detecting spammers. Liu et al. \cite{Liu2018HeterogeneousGN} proposes an edge-type weighted graph convolution network over a user-device graph to detect abusive accounts. \cite{zeng2021rlc} proposes a residual layer-GNN whereas \cite{peng2021reinforced} proposes a recursive and flexible neighborhood selection guided multi-relational GNN to detect evasive abusers. Both these techniques identify only a narrow set of abusers.

\section{User Interaction Graph}
\label{sec:user_interaction_graph}
In this section, we first discuss an analysis of users' behavior on a real dataset from Pinterest and then describe the construction of our graph framework based on these insights. Pinterest dataset includes a range of user activities over two months: Pin creation with weblinks, interaction with other users' content, user following, and other engagements.

\subsection{Spam Behavior Analysis}
\label{sec:spam_behavior_analysis}
Here, we present a behavioral analysis of spammers and non-spammers that uncovers distinctive patterns between them. Our analysis strongly motivates representing users' details and interactions in a graph setting to detect spammers.

We analyzed users' following ratio which is the fraction of users followed by a given user who later followed it back. We found the following ratio of spammers ($0.1\%$) to be significantly lower than non-spammers ($16\%$) because spammers blindly follow a large number of users who do not follow them back. This highlights differences in the behavior of individual spammers and non-spammers. Spammers also tend to collude and share resources, e.g., IP, device fingerprint, content, etc., to scale their spam activities at a controlled cost and thus, maximize gain. We analyzed IP sharing between pairs of users and observed a probability of $0.97$ for both users being spammers compared to a probability of $0.027$ for one of them being a spammer and only a probability of $0.001$ for both users being non-spammers. We also studied content sharing between pairs of users using Jaccard's metric and observed that spammers have a much higher sharing than non-spammers.% as they create content from a shared pool. 

Next, we analyzed users' interaction with weblinks using a weighted users-domain bipartite graph where the weight of an edge is the frequency of a user's interactions with a domain. We found that non-spammers interacted with twice more domains than spammers but interacted only half times that of spammers. This shows that spammers tend to highly engage with a small number of domains. We also investigated neighborhood patterns in this bipartite graph. For a user, we define the $1$-hop neighbors as users who have interacted with the same domain. We found that spammers have $5$ times more spammers in their neighborhood than non-spammers. We also computed the average shortest path between spammers and found it to be $1.4$ times more than non-spammers. This shows that spammers tend to highly engage with a small set of disconnected domains in groups. We also analyzed user-user graphs based on IP and content sharing which revealed that spammers have $6$ times and $4$ times more spammers in their ego-networks than non-spammers in these graphs respectively.

\subsection{Graph Construction}
\label{sec:graph_construction}
Here, we describe our approach to construct the graph that holistically captures users' profile details and overall behavior. We represent users as nodes and details specific to users as node attributes. We categorize entities into two groups and take a mixed approach to represent users' interactions with entities. For entities with additional information like domain, we represent these as nodes and their details as node attributes. We represent users' interactions with these entities by typed edges with attributes. Edge types distinguish users' interactions with entities by entity types. For entities with no additional information like IP, we create a typed edge with attributes between a pair of users based on their interaction with the entity type. This mixed approach simultaneously captures more behavior details while limiting the size of the graph. A schematic overview of this user interaction graph is shown in Figure~\ref{fig:model_overview}.

\section{Problem Definition}
\label{problem_defintion}
Formally, our graph is represented as a directed heterogeneous multi-graph $\mathcal{G}=\{\mathcal{V}, \mathcal{E}, \mathcal{R}, \mathcal{T}_{v}\}$ with nodes $v_i \in \mathcal{V}$ and edges $(v_i, r, v_j) \in \mathcal{E}$, where $r \in \mathcal{R}$ represents a relation and $\mathcal{T}_{v}$ represents the set of all node types. Nodes $\mathcal{V}$ in the graph consists of users $u \in \mathcal{U}$ and entities $\mathcal{O}= \{\mathcal{O}^p\}$ of different types $p \in \mathcal{T}_v\setminus u$ which the users interacted with. Additionally each node $v$ of type $t \in \mathcal{T}_{v}$ is associated with a set of features $x_v \in \mathbb{R}^{d_t}$ and each edge $e \in \mathcal{E}$ of relation $r \in \mathcal{R}$ is associated with features $x_e \in \mathbb{R}^{d_r}$.

Given the graph $\mathcal{G}$, our task is to learn a function $f(u, \mathcal{G}) \rightarrow \mathbb{R}$ to predict the spam probability of a user node $u \in \mathcal{U}$ using both its user features $x_u$ and neighborhood with edge types and attributes.

% and labeled
\section{Model}
\label{model}
Here, we describe our end-to-end learned model \modelname which consists of two main components. First, an encoder $\mathcal{ENC}$ maps a node $v$ to an embedding space $\mathbb{R}^{d}$ using its local neighborhood and features. Second, an output layer predicts the probability of spam for a user node $u$ based on its embedding $\mathcal{ENC}(u)$.

The encoder leverages rGCN~\cite{rGCN} as its building block due to its effectiveness in simultaneously capturing both the relational neighborhood structure and the local information associated with nodes. The encoder consists of several layers of convolution in the graph and is shown schematically in Figure~\ref{fig:model_overview}. Given a latent representation $h_v^{(l)}$ of a node $v \in \mathcal{V}$ in the $l$-th layer of the neural network, the encoder can be represented as:
\begin{equation}
\label{eq:layer}
h_v^{(l+1)}= \sigma \left(W_s^{(l)}h_v^{(l)} + \sum_{r \in \mathcal{R}}AGG_{r}(\mathcal{N}^r_v)   \right),
\end{equation}
where $\sigma$ is the RELU activation function, $W_s^{(l)}$ is the transformation matrix for self-loop and $\mathcal{N}^r_v$ is the set of neighbors of node $v$ for relation $r\in\mathcal{R}$. $h_v^{(0)}$ is initialized by a linear projection of node features $x_v \in \mathbb{R}^{d_t}$ of node $v$ using a node type  $t  \in \mathcal{T}_v$ specific transformation matrix $W_{t}$:
\begin{equation}
h_v^{(0)} = W_{t}x_v\,.
\end{equation}

%associated with each edge in the graph
We incorporate edge features in the per-relation neighbor aggregator function $AGG_{r}$. For each edge $e=(v, r, v')$, we obtain the edge weight $w_e^{(l)} \in \mathbb{R}$ based on its features $x_e \in \mathbb{R}^{d_r}$ using the relation-specific learned function $\phi_r^{(l)}: \mathbb{R}^{d_r} \rightarrow  \mathbb{R}$  as
\begin{equation}
    w_e^{(l)} = \phi_r^{(l)}(x_e) = \sigma \left(MLP^{(l)}_{r}(x_e) \right),
\end{equation}
where $MLP^{(l)}_{r}$ is a two-layer perceptron and $\sigma$ is the logistic sigmoid function. The aggregator incorporates these edge weights to summarize the neighbors of a node $v$ as follows,
\begin{equation}
    AGG_{r}(\mathcal{N}^r_v) = \frac{1}{|\mathcal{N}^r_v|}\sum_{v' \in \mathcal{N}^r_v} W_r^{(l)} (w_e^{(l)} h_{v'}^{(l)}), \text{ where } e = (v, r, v') \,
\end{equation}
where $W_r^{(l)}$ is the transformation matrix for relation $r \in \mathcal{R}$. Since the transformation matrix depends on the relation type, the encoder propagates latent node feature information across the edges of the graph while taking the edge type into account.

Each user node has an output layer to learn spam probability $p(u)$ defined as:
\begin{equation}
p(u) = \sigma (W h_u + b),
\end{equation}
where $W \in \mathbb{R}^{d}$ is the weight matrix, $b$ is the scalar bias, and $\sigma$ is the logistic sigmoid function.

We train \modelname end-to-end using the labeled subset of user nodes $\mathcal{U}_l \subset \mathcal{U}$ with binary cross-entropy loss as follows:
\begin{equation}
    L = - \frac{1}{|\mathcal{U}_l|} \sum_{u \in \mathcal{U}_l} y_u \log p(u)) + (1-y_u) \log (1-p(u)),
\end{equation}
where $y_u$ is the spam label for a user node $u$. 

\section{Experiments}
\label{experiments}

In this section, we present a comprehensive evaluation on a large real Pinterest dataset with multiple ablations and qualitative analysis and provide comparisons with state-of-the-art methods on a public Amazon Fraud detection dataset~\cite{dou2020enhancing} to establish the effectiveness of \modelnamenospace.

\subsection{Experiment Setup}
\textbf{Datasets.}
\textit{Pinterest Dataset}. This dataset consists of user activities on Pinterest — a visual discovery engine for exploring billions of inspirations. Users on Pinterest can create, browse, search and save (repin) Pins on boards. Each Pin has media content that may link to an external web page. Labels for users are obtained by a mix of human reviews and spam users detected by highly precise production systems both of which enforce Pinterest's community guidelines \footnote{\text{https://policy.pinterest.com/en/community-guidelines}}. 

For each user in the dataset, we collect their activities over a period of $2$ months to get good coverage of their behavior. We time-split the data into training and test to ensure that these are non-overlapping consecutive time windows of activities. We include only labels for users who were detected and actioned after the end of the time window to replicate the actual production scenario and avoid data leakage.

 Our graph, constructed as described in \ref{sec:graph_construction}, includes users' interactions with three types of entities: domain (derived from the associated Pin weblinks), IP, and content. Domains are represented as nodes in the graph along with users whereas users' interactions with IPs and content are reduced to user-user edges. We drop domains that have more than $10,000$ users interacting with them and users who have less than $10$ interactions with domains. A user node has $15$ features including locale, signup details, follow features, and frequency of changes in profile attributes. A domain node has $3$ features including spamminess score and spam label. 
 The graph consists of $4$ relations, namely, 
 \begin{enumerate}
     \item U-I-D: connecting users to domains having at least $10$ interactions, \item D-I-U: connecting domains to users having at least $10$ interactions,
     \item U-E1-U: connecting users who share the same IP at least once, and 
     \item U-E2-U: connecting users who share content at least once. 
 \end{enumerate}
 A user-domain edge has $35$ features including the number of interactions and proportion of interactions to this domain. A user-user edge has features like frequency of sharing and Jaccard's metric over the set of IPs and content between corresponding users. Table \ref{tab:data_statistics} shows the number of nodes and edges for each relation in both the train and test split of the Pinterest dataset.
\vspace{\topsep}
\textit{Amazon Fraud Dataset}. This is a subset of Amazon’s product dataset~\cite{mcauley2013amateurs}. We construct a heterogeneous graph as in~\cite{dou2020enhancing} with three relations, namely, 
\begin{enumerate}
    \item U-P-U: connecting users reviewing at least one common product,
    \item U-S-V: connecting users having at least one same star rating within one week, and
    \item U-V-U: connecting users with top $5\%$ mutual review text similarities (measured by TF-IDF) among all users.
\end{enumerate}
The number of nodes and edges for each relation is shown in Table~\ref{tab:data_statistics}. We use only $40\%$ of the nodes for training similar to other works on this dataset.
\begin{table}[tb]
\begin{center}
\resizebox{1\linewidth}{!}{
\begin{tabular}{@{}ccrrcr@{}}
\toprule
Dataset                        & Node Type       & \#Nodes & \begin{tabular}[c]{@{}l@{}}\#Labeled \\Nodes\\ (\% Spam)\end{tabular} & Relation              & \#Edges   \\ \midrule
\multirow{4}{*}{Pinterest-Train} & User   & $22.21$M     & 85,626 (63\%)                                                                  & U-I-D & $76.1$8M        \\
                               & Domain & $2.37$M      & -                                                                  & D-I-U    & $76.18$M        \\
                               & ALL    & $24.58$M     &  -                                                                   & U-E1-U    & $801.01$M       \\
                               &        &         &                                                                     & U-E2-U    & $265.14$M        \\
                               &        &         &                                                                     & ALL                   &    $1.22$B       \\ \midrule
\multirow{4}{*}{Pinterest-Test}                   & User   & $21.19$M     & 42,211 (63\%)                                                                  & U-I-D & $67.71$M        \\
                               & Domain & $2.52$M      & -                                                                  & D-I-U    & $67.71$M        \\
                               &        &         &                                                                     & U-E1-U    & $923.99$M       \\
                               &        &         &                                                                     & U-E2-U    & $341.32$M        \\
                               &        &         &                                                                     & ALL                   & $1.40$B        \\ \midrule
\multirow{4}{*}{Amazon-Fraud}                   & User   & $11,944$  & $11,944$ ($9.5\%$)                                                     & U-P-U                 & $175,608$   \\
                               &        &         &                                                                     & U-S-U                 & $3,566,479$ \\
                               &        &         &                                                                     & U-V-U                 & $1,036,737$ \\
                               &        &         &                                                                     & ALL                   & $4,398,392$ \\ \bottomrule
\end{tabular}
}
\caption{Statistics of nodes and edges per relation for various graph datasets used for empirical evaluations. Column $4$ shows the number of labeled user nodes in each dataset along with the percentage of spammers among them.}
\vspace{-2em}
 \label{tab:data_statistics}
 \end{center}
\end{table}
\\
\\
\textbf{Baselines.} For ablation evaluations, we use a no-graph model using only user features and a \modelname variant without edge features to show the importance of incorporating interactions and edge features in the model. For comparison with the state-of-the-art methods, we use RLC-GNN~\cite{zeng2021rlc} and RioGNN~\cite{peng2021reinforced} (a multi-layer extension of the CARE-GNN~\cite{dou2020enhancing} model) both of which leverage reinforcement learning (RL) to select an optimal number of neighbors to address spammer evasiveness.\\\\
\textbf{Experiment Settings.}
Based on multiple hyperprameter tuning runs, we select node embedding size ($d$) of $256$, batch size of $512$, $2$ network layers, a learning rate of $9.5\times 10^5$, LayerNorm~\cite{ba2016layer} after each GCN layer, and L2 regularization weight of $0.0001$ for all models. To improve the training efficiency on a large-scale graph, we employ neighbor sampling~\cite{hamilton2017inductive} with a fanout factor of $50$ for each relation for both layers to train all graph neural network models. We implement all models in Pytorch~\cite{NEURIPS2019_9015} using DGL~\cite{wang2019dgl} and train them on $2$ GPUs with Adam optimizer. We train all models for a maximum of $2,000$ steps and use the validation set (a subset of nodes in the training graph) for early stopping.\\\\
\textbf{Evaluation Metrics.} A key requirement of any spam detection system is to have a low \textit{false-positive rate (FPR)} while having a high recall in identifying spammers to ensure minimal impact on genuine users. Hence we use recall at an FPR of $1\%$ as our evaluation criteria to compare different models. We also report the normalized area under the ROC curve truncated at $1\%$ FPR to analyze models' performance in regions with low-false positive rates.

\subsection{Results}

We report test set Recall@FPR1 and AUROC@FPR1 for \modelname and other baselines on Pinterest dataset in Table~\ref{tab:results-private}. Our model outperforms all baselines by achieving $80.06\%$ Recall@FPR1 and $69.52\%$ AUROC@FPR1. \modelname has $40\%$ higher recall over no-graph user features only baseline demonstrating the effectiveness of graph in identifying spam behavior. We also see that learning edge weights from attributes help \modelname get $3\%$ improvement in recall at $1\%$ FPR over its variant with no edge features.

\begin{table}[tb]
\begin{tabular}{@{}lcc@{}}
\toprule
Model                   & Recall@FPR1($\%$) & ROCAUC@FPR1($\%$) \\ \midrule
\modelname                   & $80.06$        & $69.52$        \\
\modelname w/o edge features & $77.95$       & $64.85$        \\
User only baseline         & $57.31$        & $33.41$        \\ \bottomrule
\end{tabular}
\caption{Performance of \modelname and baseline methods on the Pinterest test dataset.}
\vspace{-1.6em}
\label{tab:results-private}
\end{table}

We show comparisons with the state-of-the-art methods on the Amazon dataset in Table~\ref{tab:results-amazon}. \modelname outperforms \textit{RioGNN} and has comparative performance with \textit{RLC-GNN} despite being much simpler in architecture and not using expansive RL techniques for neighbor selection. Both \textit{RioGNN} and \textit{RLC-GNN} compute similarity of a node with all its neighbors to perform neighbor selection and hence cannot scale to web-scale graphs. This shows the superiority of \modelname in terms of scalability and performance over state-of-the-art methods in learning a good representation of users by exploiting their rich interactions on the platform.
\begin{table}[bt]
\begin{tabular}{@{}lcc@{}}
\toprule
Model   & Recall@FPR1($\%$) & ROCAUC($\%$) \\ \midrule
\modelname   & $73.84$       & $96.69$  \\
RLC-GNN~\cite{zeng2021rlc} & -           & $97.48$  \\
RioGNN~\cite{peng2021reinforced} & -           & $96.19$  \\ \bottomrule
\end{tabular}
\caption{Comparison of \modelname with the state-of-the-art methods on Amazon Fraud dataset.}
\vspace{-2em}
\label{tab:results-amazon}
\end{table}
\begin{figure}
    \centering
    \includegraphics[width=\linewidth]{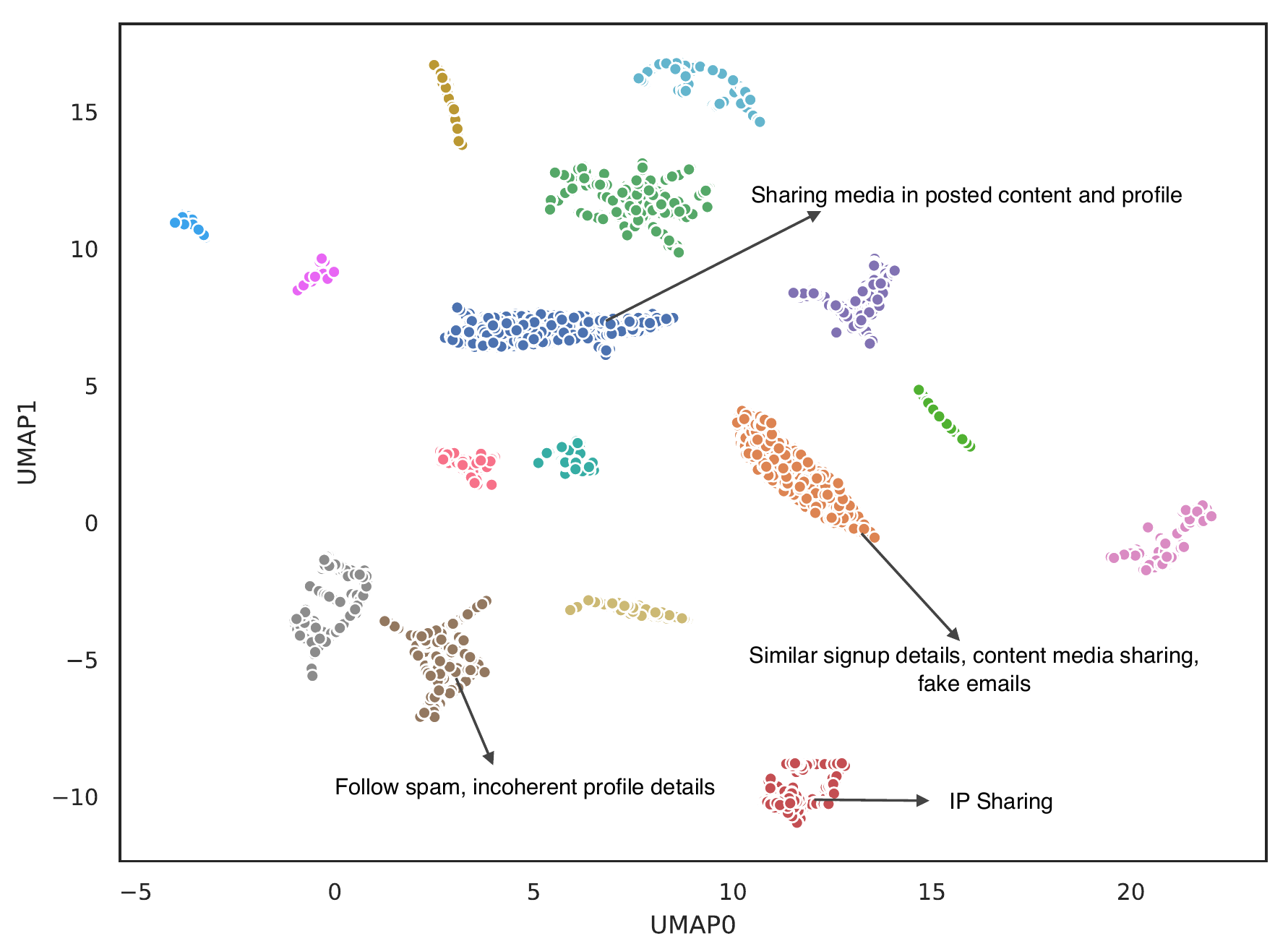}
    \caption{UMAP visualization of clusters of users detected as spam by \modelnamenospace. Some clusters are annotated with the abusive behavior exhibited by the users in the cluster. Users exhibiting similar abuse behavior are closer to each other in the embedding space demonstrating that \modelname can extract a wide range of abusive behavior and learn meaningful representations of them.}
    \label{fig:visualization}
\end{figure}

To further understand the learned representations of the users, we analyzed embeddings of spammers identified by \modelnamenospace. We use DBSCAN~\cite{ester1996density} to cluster the users by their embeddings and use UMAP~\cite{mcinnes2018umap} to visualize the clusters as shown in Figure~\ref{fig:visualization}. We sampled a few users from each of the big clusters and analyzed their spam behavior. We observed that each cluster corresponds to a different abusive behavior as annotated in the figure. Further, users exhibiting similar abuse behavior are closer to each other in this space. This shows that \modelname can extract a wide range of abusive behavior and learn meaningful representations of them.

\section{Conclusion}
\label{conclusion}
Addressing spam on a web-scale platform is very challenging because of its scale and a wide range of abusive behavior of spammers. In this paper, we proposed a new heterogeneous graph framework that holistically captures users' behavior and facilities learning a powerful graph convolution network model to identify spammers with high coverage. Our model considers additional graph structures like edge types and attributes for learning which boosts its performance. We provide strong empirical results on two real datasets to show its superior performance over alternative and ablation methods. Our results also show that the method achieves a very high recall with a small false-positive rate and hence, can be deployed in real production to effectively address spam.

\begin{acks}
We thank Rundong Liu, Omkar Panhalkar, Yuanfang Song, and Dennis Horte for their valuable inputs. We also thank Maisy Samuelson for her review and feedback.
\end{acks}
%%
%% The next two lines define the bibliography style to be used, and
%% the bibliography file.
\balance
\bibliographystyle{ACM-Reference-Format}
\bibliography{userspam}

% \appendix
% \section{Pinterest spam behavior analysis}
% \label{data-analysis}
% \input{chapters/analysis}
%%
%% If your work has an appendix, this is the place to put it.
%\appendix
%
%\section{Research Methods}
%\subsection{Part One}
%\subsection{Part Two}
%
%\section{Online Resources}
\end{document}